\documentclass[letterpaper]{article} 
\usepackage{aaai24}  
\usepackage{times}  
\usepackage{helvet}  
\usepackage{courier}  
\usepackage[hyphens]{url}  
\usepackage{graphicx} 
\usepackage{lscape}
\usepackage{adjustbox} 
\usepackage{array} 
\usepackage[numbers]{natbib} 
\usepackage{natbib}  
\usepackage{caption} 
\usepackage{algorithm}
\usepackage{algorithmic}

\usepackage{newfloat}
\usepackage{listings}
\usepackage{xcolor}
\usepackage[hidelinks]{hyperref}
\usepackage{booktabs}
\DeclareCaptionStyle{ruled}{labelfont=normalfont,labelsep=colon,strut=off} 
\floatstyle{ruled}
\newfloat{listing}{tb}{lst}{}
\floatname{listing}{Listing}
\title{Human-Modeling in Sequential Decision-Making:\\
An Analysis through the Lens of Human-Aware AI}

\author{
Silvia Tulli {\rm 1}
Stylianos Loukas Vasileiou {\rm 2}\and
Sarath Sreedharan {\rm 3}
}
\author {
    Silvia Tulli\textsuperscript{\rm 1}\footnote{Contact Author},
    Stylianos Loukas Vasileiou \textsuperscript{\rm 2},
    Sarath Sreedharan \textsuperscript{\rm 3}
}
\affiliations{
\textsuperscript{\rm 1}Institute of Intelligent Systems and Robotics (ISIR) - CNRS - INSERM - Sorbonne University\\
\textsuperscript{\rm 2}McKelvey School of Engineering at Washington University in St. Louis\\
\textsuperscript{\rm 3}Department of Computer Science at Colorado State University\\
\texttt{tulli@isir.upmc.fr},
\texttt{v.stylianos@wustl.edu},
\texttt{sarath.sreedharan@colostate.edu}}

\newcommand{\Shortcite}[1] {\citeauthor{#1}~\shortcite{#1}}

\newboolean{includeMemo}
\setboolean{includeMemo}{true} 

\newcommand{\memo}[1]{\ifthenelse{\boolean{includeMemo}}{\todo[inline,caption={},color=green!20!]{#1}}}
\newcommand{\memob}[1]{\ifthenelse{\boolean{includeMemo}}{\todo[inline,caption={},color=blue!20!]{#1}}}

\begin{document}
\maketitle

\begin{abstract}
``Human-aware'' has become a popular keyword used to describe a particular class of AI systems that are designed to work and interact with humans. 
While there exists a surprising level of consistency among the works that use the label human-aware, the term itself mostly remains poorly understood.
In this work, we retroactively try to provide an account of what constitutes a human-aware AI system.
We see that human-aware AI is a design oriented paradigm, one that focuses on the need for modeling the humans it may interact with.
Additionally, we see that this paradigm offers us intuitive dimensions to understand and categorize the kinds of interactions these systems might have with humans.
We show the pedagogical value of these dimensions by using them as a tool to understand and review the current landscape of work related to human-AI systems that purport some form of human modeling.
To fit the scope of a workshop paper, we specifically narrowed our review to papers that deal with sequential decision-making and were published in a major AI conference in the last three years.
Our analysis helps identify the space of potential research problems that are currently being overlooked. 
We perform additional analysis on the degree to which these works make explicit reference to results from social science and whether they actually perform user-studies to validate their systems. We also provide an accounting of the various AI methods used by these works.

\end{abstract}

\section{Introduction}

Artificial Intelligence (AI) is currently undergoing a transformational moment, signaling a paradigm shift in its application and perception. There is an escalating optimism surrounding AI-based systems and their potential to significantly enhance the lives of everyday users. This optimism is not just a theoretical construct but has also fostered an intense interest in the development of AI systems that are adept at collaborating with and assisting humans in meaningful ways.

As with any rapidly evolving research domain, this interest has spawned a proliferation of varied research clusters, each with its unique focus. A brief survey of the landscape in human-AI interaction research uncovers a plethora of terms that researchers employ to define their work. Prominent among these are \textit{human-centered AI}, \textit{human-compatible AI}, \textit{human-in-the-loop AI}, and \textit{human-aware AI}. To a novice in the field, these terms might appear bewildering, each suggesting subtly different goals, methodologies, or design principles. Some of these terminologies have become preferred nomenclatures within specific research communities, while others outline distinct research objectives or design paradigms. 



\textit{Human-centered AI} (or human-centric AI) places humans at the core of the design process, primarily focusing on enhancing the human user experience. A system adopting a human-centered approach may not necessarily be human-aware. In this context, being human-centric implies a focus on enhancing user experience by accounting for human factors, such as preferences, needs, and values, in engineering and design, without necessarily incorporating a detailed modeling of human behavior \cite{chetouani2023humancentred}. This contrasts with \textit{human-compatible AI}, which is more about the type of problems being addressed, especially those related to AI safety and ethical considerations \cite{russell2019humancompatible}. The term \textit{human-in-the-loop AI} has evolved over time; earlier works emphasized more explicit human intervention in the decision-making process \cite{retzlaff2024HumanintheLoopRL}, but more recent interpretations often relate to machine learning scenarios where humans play a more collaborative role in the learning process of the AI system \cite{mosqueira2023human}.

Our paper focuses on the last category, namely \textit{human-aware AI}. To the best of our knowledge, this term has been used by multiple research groups and communities in a surprisingly consistent manner, making it a particularly intriguing area of study. The term ``human-aware'' first entered the AI lexicon consistently with the work of \cite{alami2005}. Even in these early stages, many of the characteristic features of subsequent research using this term were evident. This foundational work demonstrated how an AI agent, such as a robot, needs to model human behaviors and infer their intentions, particularly goals, to facilitate more fluid interactions. The connection between this modeling, mental models, and the theory of mind was noted by \citet{devin2016implemented}, and this concept has since been expanded in later works (cf. \cite{chakraborti2018human}) to include a broader range of models. Subsequent research has utilized this premise to offer formal accounts of various phenomena in AI, including explainability \cite{sreedharan2021foundations,vasileiou2022jair}, trust \cite{zahedi2023mental}, and value alignment \cite{mechergui2024goal}, as well as more comprehensive models of human-AI interaction (cf. \cite{kambhampati2020challenges}).

The aim of this paper is not merely to present another formal account of some aspect of human-AI interaction. Rather, our goal is to step back and provide a general account of what it means for an AI approach to be human-aware. We then intend to apply this framework to analyze a wide array of current AI works that focus on some form of human interaction, assessing their alignment with the human-aware AI paradigm.

We assert that, unlike other terms discussed earlier, human-aware AI systems are characterized by two interrelated yet distinct features:\footnote{These two features together imply that all human-aware AI systems are inherently multi-agent systems.}

\begin{itemize}
    \item \textit{Acknowledgment of Human Interaction} (F\textsubscript{1}): This entails an explicit acknowledgement that the AI system will, at some point in its lifecycle, interact with humans.

    \item \textit{ Design Consideration for Human Interaction}  (F\textsubscript{2}): This feature goes beyond mere acknowledgment, requiring that the AI system's design 
    considers human modeling to account for the anticipated human interaction.
    
\end{itemize}

Taking a closer look at F\textsubscript{1}, we observe that virtually all AI systems, including those as remote as the Mars rovers, interact with humans in some capacity. However, our focus is on whether this interaction is explicitly acknowledged and integrated into the system's design. Many AI systems are initially conceived as single-agent systems, with human interaction considerations often incorporated as an afterthought. A relevant example is powerful Reinforcement Learning (RL) systems like Alphafold \cite{jumper2021highly}, which are designed to solve specific problems (e.g., protein folding) rather than focusing on end-user interaction.

In contrast, F\textsubscript{2} mandates that for an AI system to qualify as human-aware, its design must be influenced by the necessity of human interaction. This feature presupposes F\textsubscript{1}, but distinguishing between the two adds clarity. A system designer might be aware that human interaction will occur but may deem explicit human modeling unnecessary for certain use cases. We argue that such a system still qualifies as a human-aware AI system, as the potential for human interaction was considered during its design phase.

It is crucial to note that being a human-aware AI system does not automatically imply effectiveness in this role. Echoing recent discussions in fields like explainable AI (XAI) \cite{gunning2019darpa}, we propose that the most reliable method to evaluate a human-aware AI system's efficacy is through human subject studies. Thus, an effective human-aware AI system is one that demonstrates practical utility in real-world human interactions.

The rest of the paper is structured as follows: We begin with a discussion of the various roles that humans could play in a human-aware system. Next, we review recent papers from major AI conferences to evaluate whether they align with the features of human-aware AI as outlined above. This analysis will categorize these works based on the roles played by humans, the influence of human interaction on system design, and the degree of utilization of human interactions.

\section{Humans and Human-Aware AI System}

\begin{figure}
    \centering
    \includegraphics[scale=0.1]{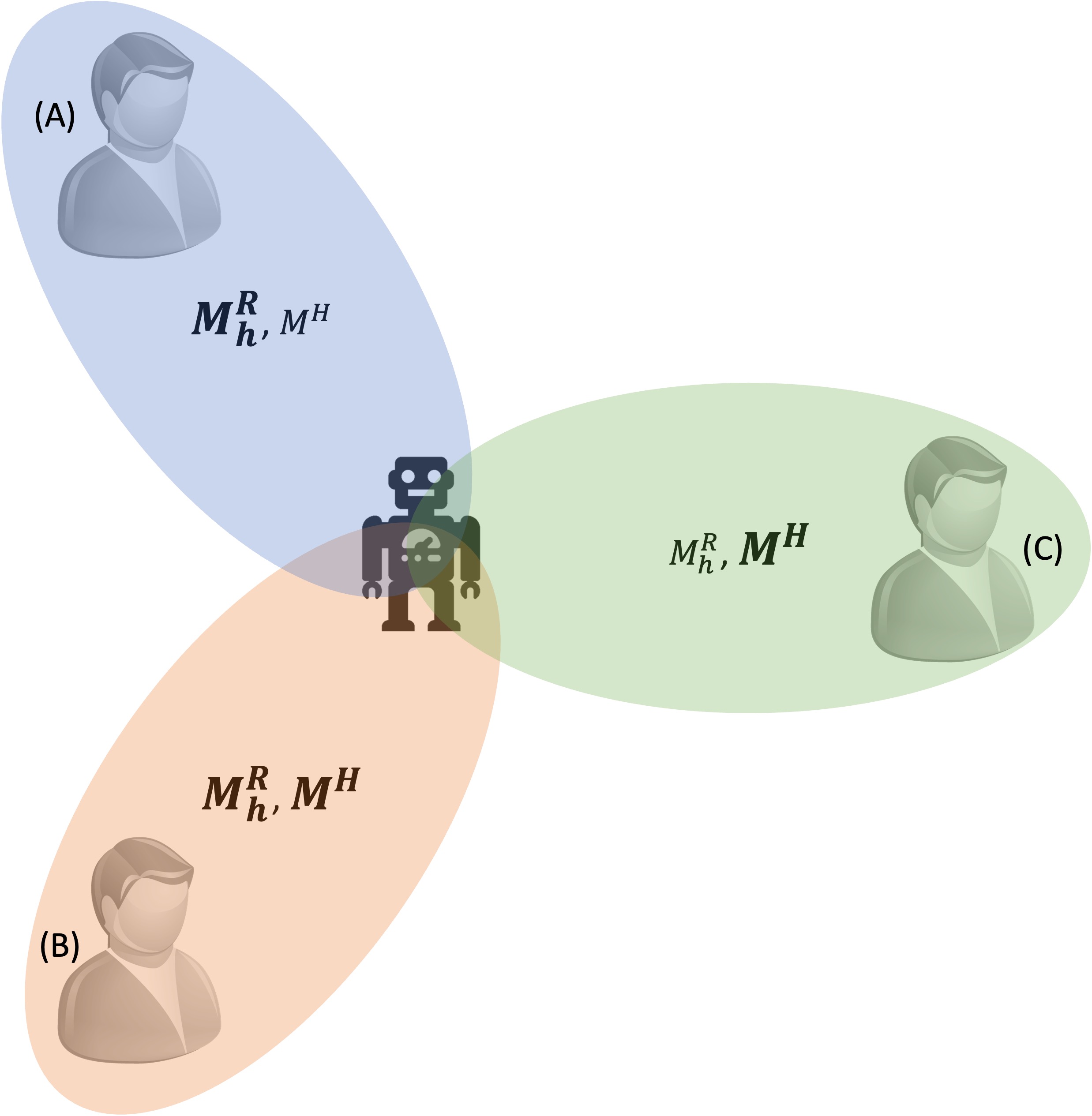}
    \caption{A visualization of the three modes in which the human and AI systems could interact and the models that would come into play in such interactions (the most important one is bolded and put in a slightly larger font). In the context of teaming or collaboration, the three categories correspond to (A) Supervisor/Teacher (B) Teammate (C) End-User.}
    \label{fig:categories}
\end{figure}

We now examine the specifics of how humans may interact with AI systems. Figure \ref{fig:categories} offers an overview of the broad categories into which human roles and interactions with AI might fall. As previously discussed, an explicit model of human behavior is not a prerequisite for a method to be categorized as human-aware AI. Nonetheless, models of human beliefs and capabilities serve as useful tools for understanding and categorizing different modes of interaction between humans and AI.

We approach human-aware AI from the perspective of multi-agent systems, where both humans and AI are considered agents. These agents' actions or decisions are predicated on internal models, which capture their knowledge of the task and decision-making processes. It is important to clarify that our use of ``model'' here is a heuristic tool for discussing agent states and behaviors, devoid of any assertions about the actual cognitive processes of humans or AI systems. This interpretation accommodates even model-free decision-making paradigms by assuming a basic model comprising reflexive rules.

While there are existing formulations that consider a lot more types of mental models (cf. \cite{zahedi2022mental}), for the purposes of this paper, we focus on two primary models:
\begin{itemize}
    \item $\mathcal{M}^H$: The human model encompasses the human's knowledge about the world state, its capabilities, and current goals/preferences. 

    \item $\mathcal{M}^R_h$: The model of the human's perception of the AI agent, detailing the human's knowledge about the agent's understanding of the world, its capabilities, and goals.
\end{itemize}

These models underpin three principal categories of human roles in human-AI interactions:

\begin{enumerate}
    \item \textit{Supervisor/Teacher}: In this role, the human oversees the agent's operations, providing feedback or guidance. The critical model here is $\mathcal{M}^R_h$, as the supervisor must have an understanding of the agent's knowledge to offer relevant feedback. However, $\mathcal{M}^H$ also plays a role, as effective supervision requires an awareness of human capabilities and goals.

    \item \textit{Teammate}: Here, the human actively collaborates with the agent to achieve a mutual goal. Both models are crucially important, enabling the human to anticipate the agent's decisions ($\mathcal{M}^R_h$) and plan their own actions ($\mathcal{M}^H$). Similarly, the agent requires an understanding of the human's actions and expectations.

    \item \textit{End-User}: As an end-user, the human primarily interacts with the AI system as a beneficiary of its services. The predominant model in this scenario is $\mathcal{M}^H$, as the agent's goal is to assist the human. The human's perception of the agent ($\mathcal{M}^R_h$) also bears significance to the extent that it influences the human's expectations and utilization of the system.

\end{enumerate}

It is important to note that these roles are not mutually exclusive; a single individual may assume multiple roles within the same or different interactions. Furthermore, these categories can be extended to multiple user scenarios and even to adversarial contexts, such as:

\begin{enumerate}
    \item \textit{Attacker}: Corresponding to the Supervisor/Teacher, with a focus on undermining the agent (primary model: $\mathcal{M}^R_h$).

    \item \textit{Rival}: Analogous to the Teammate, competing with the agent for resources or goals (both models are significant).

    \item \textit{Target}: Similar to the End-User, being the focus of the agent's adversarial actions (primary model: $\mathcal{M}^H$).

\end{enumerate}

\noindent This paper, however, will concentrate on collaborative rather than adversarial aspects of human-AI interactions.

The works in this space, tends to characterize the level of human modeling using the following three dimensions \cite{sreedharan2023human}:
\begin{enumerate}
    \item \textit{Knowledge State}: This corresponds to human knowledge or belief, i.e., the contents of the specific models mentioned earlier.
    \item \textit{Inferential Capability}: This corresponds to how the human may make use of the given model to come up with plans or decisions. Generally, humans are widely accepted to be bounded rational agents \cite{noisyrat}, even though they are not always modeled as such. Also, it's worth considering that in the case of $\mathcal{M}^R_h$, the system would need to capture the inferential capability the human ascribes to the system itself.
    \item \textit{Vocabulary}: This corresponds to the terms in which the human represents and reasons about the task, which in turn could influence their decisions and interactions. It is worth noting that the same task could in theory be captured using different terms.
\end{enumerate}
\section{Methodology}

In this section, we describe our methodology for evaluating recent works with respect to the human-aware AI criteria outlined in the previous section. 

We opted for a review of papers published between 2020 and 2023 and targeted major AI conferences\footnote{AAAI, IJCAI, ECAI, and ICAPS}.
In particular, we opted for a keyword-based search within the Semantic Scholar database, using Semantic Scholar API\footnote{\href{https://api.semanticscholar.org/api-docs/}{https://www.semanticscholar.org/product/api}} to gather and analyze literature.
Our use of the API and fixed keywords also provide this approach with a level of reproducibility, that is usually missing from most surveys.
While this method does not provide us with an exhaustive characterization of the entire landscape of AI methods, it does give us an overview of, at the very least, the recent trends in the field.


We filtered papers by searching for content including the term ``human", as well as related terms identified from the search relevance algorithm of Semantic Scholar, and ``compatible" or ``aware" or ``Theory of Mind (ToM)" or ``modeling" and ``plan''.
This was done to specifically identify studies that are directly relevant to human users. Additionally, we focused on work considering sequential decision-making processes, a critical aspect of human-AI interaction. Table~\ref{tab:inclusion_criteria} shows our inclusion criteria and search strings used on the Semantic Scholar API. We narrowed our focus to a more recent three-year period within our initial ten-year range.

Our filtering approach resulted in an initial pool $312$ papers. We conducted an exploratory analysis of these papers to gain an overview and identify any general trends. 
Then, we conducted a manual filtration of this list, further eliminating unrelated work by assessing relevance based on titles, abstracts, and thorough readings of the full texts. We identified $66$ relevant papers. To ensure the reliability of our selection, after this initial screening, we redistributed the papers among the authors. This ensured that each paper was deemed relevant by at least two authors. For papers with differing decisions, we discussed their relevance and made a joint decision regarding the paper’s inclusion. This left us with a final list of $46$ related papers.


\begin{table*}[t]
\resizebox{\textwidth}{!}{%
\begin{tabular}{@{}lllll@{}}
\toprule
Topic & Description & Search Term &  &  \\ \midrule
\begin{tabular}[c]{@{}l@{}}Human Involvement \end{tabular} & \begin{tabular}[c]{@{}l@{}}Exclude papers without human\\ involvement or considerations\\ (e.g., position\\ papers).\end{tabular} & \textit{human} AND \textit{compatible} OR \textit{aware} OR \textit{ToM} OR \textit{modeling} &  &  \\  \\

\begin{tabular}[c]{@{}l@{}}Sequential Decision \\ Making \end{tabular}  & \begin{tabular}[c]{@{}l@{}}Exclude papers that do not use\\ an agent.\end{tabular} & \textit{plan} &  &  \\  \\
Recency & Only consider the past 3 years. & \textit{range}(2020, 2023) &  &  \\ \\
Subject Area & \begin{tabular}[c]{@{}l@{}}Only consider papers from\\ Computer Science.\end{tabular} & \textit{Computer Science} &  &  \\ \bottomrule
\end{tabular}%
}
\caption{Inclusion Criteria and Search Strings used in the Semantics Scholar API.}
\label{tab:inclusion_criteria}
\end{table*}

\section{Human Assumptions}
In our first evaluation, we looked at some of the implicit and explicit assumptions concerning humans.
While there is a shared understanding across these papers about the complexity of human behavior and the value of human input in AI systems, they differ in their focus on the nature of human emotions, the specificity of interaction contexts, the degree of human involvement, and the types of contributions humans are expected to make. \\

Firstly, there is a recognition of \textit{knowledge asymmetry} between humans and AI agents regarding capabilities and preferences \cite{Sreedharan2021AUB}. Additionally, the research acknowledges that humans often plan individually while also considering parallel planning with other humans \cite{Czechowski2021}. Moreover, human decision-making involves various uncertainties and anxieties about future outcomes \cite{Vanhe2022AnxietySensitivePF}. This uncertainty extends to beliefs about AI agents \cite{Sreedharan2021AUB}.

Human conversations are viewed as goal-oriented and guided by multiple small goals or a global goal \cite{Ni2022HiTKGTG}. Furthermore, human driving behavior is recognized as diverse and influenced by individual priorities and motivations \cite{Sarkar2021SolutionCI} \cite{Sarkar2021GeneralizedDC}.

The need for \textit{explanations} in human-AI interaction is emphasized in several papers \cite{sreedharan2020expectation,vasileiou2021exploiting,Kumar_Vasileiou_Bancilhon_Ottley_Yeoh_2022,ijcai2023p795,vasileioua2023please,Selvey2023FormalEO}. Additionally, personalized explanations are deemed essential \cite{vasileioua2023please}. Effective dialogue with humans also requires topic management \cite{Xu_Wang_Niu_Wu_Che_2020}.

Moreover, the assumption that diverse plans can be used as a proxy to cover unknown human preferences or that human preferences may be private or complex is highlighted \cite{Ghasemi_Scope-Crafts_Zhao_Topcu_2021}.

One of the important takeaways from these assumptions is the fact that they act as a way to incorporate information about human models without dealing with the overhead of performing explicit modeling. While baking in fixed assumptions about humans could be limiting from a modeling point of view, we do see them being effective in the scenarios considered by the papers.

\section{Human Models}
An overwhelming percentage of papers look at modeling the human's \textit{knowledge state}, with the majority of those papers focusing on using (or learning) $\mathcal{M}_h$. However, many of these works focus on different types of model representation and components. Several papers \cite{Ni2022HiTKGTG,Zhang_Kemp_Lipovetzky_2023,Katz_Ram_Sohrabi_Udrea_2020,Sarkar2021SolutionCI} highlighted the importance of goal-oriented interactions in human-AI systems, which could be formalized using hierarchical or goal-based models.
There were also papers that focused on modeling human preferences \cite{Xu_Wang_Niu_Wu_Che_2020,Ghasemi_Scope-Crafts_Zhao_Topcu_2021} and discussing preference models, which could be formalized using utility functions and learned using preference elicitation techniques. \Shortcite{wang2023} and \Shortcite{ijcai2021p0577} explore approaches for learning the human model from human interactions, which could involve techniques such as reinforcement learning with human feedback or imitation learning.

A few papers addressed the modeling of human \textit{inferential capabilities}. 
\Shortcite{Sreedharan2021AUB} and \Shortcite{Amado_Meneguzzi_2020} employ Bayesian models to capture human inference processes, while the work of \Shortcite{Zhang_Kemp_Lipovetzky_2023} and \Shortcite{DePeuter_Kaski_2023} address temporal aspects of human behavior and intention recognition.

Finally, the human \textit{vocabulary} modeling was the least represented among the dimensions, with just a few exceptions \cite{vasileioua2023please,Kumar_Vasileiou_Bancilhon_Ottley_Yeoh_2022}.


\section{Priori/Posteriori Inclusion of the Human}
Next, we looked at whether the human considerations were purely taken during the design/decision-making process or whether the system allowed for the humans to directly provide feedback and/or interact with it during the operation of the system. We refer to the former as \textit{priori} inclusion of humans and the latter as \textit{posterior} inclusion.

Examples of priori methods involve those that utilize human input or data as part of the planning or learning process (for example \cite{Zhang_Kemp_Lipovetzky_2023}).
As a counterpart, examples of posteriori methods include those involve humans after initial planning or decision-making stages, incorporating human feedback or interaction to refine or adjust the system's behavior \cite{Bara2023TowardsCP}.

While not explicitly mentioned, some approaches may indirectly address value alignment through human-centric design or by considering human feedback in the planning process \cite{Zheng2021ObjectiveawareTS, Xu_Wang_Niu_Wu_Che_2020, Ghasemi_Scope-Crafts_Zhao_Topcu_2021}. These works have discussed the integration of human preferences. \citet{Xu_Wang_Niu_Wu_Che_2020} incorporates human-inspired strategies to ensure coherent and user-interest-aligned dialogues in open-domain conversation generation. This reflects a priori inclusion of human considerations without direct feedback during operation. \citet{Ghasemi_Scope-Crafts_Zhao_Topcu_2021} introduced a diverse stochastic planning approach to generate varied plans that account for unknown or complex human preferences, considering human factors a priori. \citet{Zheng2021ObjectiveawareTS} embedded human roles and preferences into the system's design from the outset, with indirect mechanisms for incorporating human feedback iteratively.

Explanation generation emerges as a prominent theme in several works. While some approaches focus explicitly on generating formal explanations of AI decisions \cite{Selvey2023FormalEO} (hence posteriori method), others implicitly aim to make plans or actions more understandable to human end-users through self-explanatory plans or interpretability measures \cite{netanyahu2021phase} (hence priori methods).

Within explanation generation, one method we found to be quite popular is that of model reconciliation\cite{ijcai2021p571,vasileiou2021exploiting,Kumar_Vasileiou_Bancilhon_Ottley_Yeoh_2022,ijcai2023p795,vasileioua2023please}.

\section{Role of the Human}

The diverse roles humans play in AI systems, from passive data providers to active decision-makers, vary across research work. In many cases, humans serve as \textit{end-users}, benefiting from the outcomes or decisions generated by AI systems without directly influencing any decisions in real-time \cite{ijcai2020p12,Jiang2023ContinuousTG,Illanes_Yan_ToroIcarte_McIlraith_2020,liu2021game}. In other instances, humans take on more active roles, such as \textit{supervisors} providing guidance or feedback to AI systems \cite{jakubik2022designing,ijcai2021p0577,wang2023} or \textit{teammates} collaborating with AI agents \cite{Bara2023TowardsCP,Zhang2023AdaptationAC}. Works \cite{Ni2022HiTKGTG,Cai_Li_Huang_Yang_2021,netanyahu2021phase,Pingen2022TalkingTD} have also highlighted the evolving role of humans in AI systems, where they serve both as supervisors and end-users. These examples demonstrate a shift towards more interactive and collaborative AI systems that integrate human feedback and guidance into the decision-making process.

\section{Social Science Theories}
In general, it is heartening to see that more works acknowledge the importance of incorporating social science perspectives into AI research, recognizing that human behavior, cognition, and societal dynamics play crucial roles in the development and deployment of AI systems \cite{ijcai2020p736,Vanhe2022AnxietySensitivePF,Bara2023TowardsCP,Xu_Wang_Niu_Wu_Che_2020,Wang2020Intention2BasketAN,Peuter2022ZeroShotAI,Pang_Parks_Breazeal_Abelson_2023,Zhang_Kemp_Lipovetzky_2023,kumar2022vizxp,vasileioua2023please,ijcai2023p795}.

Among the papers mentioning social sciences theories, there are some based in the concept of \textit{theory of mind} \cite{netanyahu2021phase,ijcai2023p795,Zhang_Kemp_Lipovetzky_2023,Kumar_Vasileiou_Bancilhon_Ottley_Yeoh_2022}. The theory of mind involves the ability to attribute mental states to oneself and others, enabling individuals to understand and predict behavior based on inferred beliefs, desires, and intentions \cite{PremackWoodruff1978}.
Beyond the \textit{theory of mind}, other social science theories mentioned include: \textit{relevance theory} and \textit{population movement and settlement patterns} \cite{ijcai2020p736,vasileioua2023please}. \textit{Relevance theory} is a cognitive science theory that seeks to explain how utterances are interpreted \cite{Sperber_Wilson_1995}.
\textit{Population movement and settlement patterns} are focal points in human geography, demography, and urban planning, supported by various social science theories.

It is worth mentioning that while several remaining papers do not explicitly reference or integrate social science theories into their research, this does not necessarily imply a lack of consideration for human factors or societal implications. However, it does raise questions about the depth of understanding of these aspects.

\section{Evaluation Methods and Metrics}

We reviewed both \textit{quantitative} and \textit{qualitative} measures employed the relevant papers. Moreover, we aimed to evaluate the importance of incorporating \textit{user studies} and baseline comparisons in the assessment methodologies. 

Quantitative measures such as runtime \cite{sreedharan2020expectation}, mean reward \cite{Czechowski2021}, prediction accuracy \cite{Levy2022UnderstandingNL}, precision, and recall \cite{Sreedharan2021AUB} offer objective benchmarks for assessing the performance of systems across different tasks and domains. Some work prioritizes robustness metrics, aimed at evaluating the system's resilience against adversarial attacks, noise, or uncertainties in real-world scenarios. For instance, \cite{Killian2023RobustPO} use \textit{max regret} to quantify the worst-case performance deviation from the optimal outcome, providing a measure of robustness against unforeseen circumstances. Similarly, in autonomous driving contexts, evaluation criteria such as success rate and runtime, as seen in \cite{Zhang2023AdaptationAC,Zhang_Zhang_Zhang_Wang_Lu_Hong_2020} reflect the system's ability to adapt and make decisions in dynamic environments. 

Only a handful of papers surveyed opt to conduct user studies \cite{Ni2022HiTKGTG,netanyahu2021phase,Kumar_Vasileiou_Bancilhon_Ottley_Yeoh_2022,vasileioua2023please,ijcai2023p795,Zhang_Kemp_Lipovetzky_2023,Xu_Wang_Niu_Wu_Che_2020, Pang_Parks_Breazeal_Abelson_2023}. \citet{Ni2022HiTKGTG} focus on the system's ability to facilitate goal-oriented conversations through multi-hierarchy learning. \citet{netanyahu2021phase} employs evaluation metrics like Average Displacement Error (ADE) and Final Displacement Error (FDE) to measure the accuracy of predictions in physically-grounded abstract social events. These metrics provide insights into the system's performance in understanding and predicting human behavior, useful for applications requiring social interaction and perception. Further, \citet{Kumar_Vasileiou_Bancilhon_Ottley_Yeoh_2022} use metrics such as correction ratio and comprehension score to evaluate the effectiveness of their visualization techniques in conveying explanations to human users. Moreover, user-centric evaluation metrics, such as coherence (intra/inter-topic) and task completion time, are employed by \cite{Zhang_Kemp_Lipovetzky_2023} and \cite{vasileioua2023please} to assess the human user experience and task efficiency of AI systems.

\section{AI Methods and Learning Paradigms}

We finally examined the general AI methods and learning paradigms used across the relevant papers. Many papers use supervised learning for tasks such as understanding natural language \cite{Levy2022UnderstandingNL}, recognizing goals \cite{Sarkar2021SolutionCI}, and generating human-like dialogues \cite{Ni2022HiTKGTG}. Some researchers, such as \cite{netanyahu2021phase}, employ mixed methods for identifying and understanding social interactions \cite{netanyahu2021phase}. They combine supervised learning techniques with reinforcement learning to parameterize learning nodes with learned policies. Additionally, they incorporate imitation learning methods to acquire behavior trees from human demonstration. Planning-based approaches are prevalent, particularly in tasks involving decision-making and action generation. Classical planning techniques are used in various papers \cite{sreedharan2020expectation,Sreedharan2021AUB,Selvey2023FormalEO}. Additionally, more specialized planning methods such as hierarchical planning \cite{netanyahu2021phase} and behavior tree expansion algorithms \cite{Ghasemi_Scope-Crafts_Zhao_Topcu_2021} are employed in specific domains.


\section{Takeaways}
One of our first takeaways from the survey was that human-aware AI proved to be a surprisingly robust tool for analyzing the landscape of papers. At a first glance, human-aware AI might seem like a limited framework to be used as an analysis tool, especially given the fact that most papers do not maintain and manipulate explicit representations of human mental models. On the other hand, works that acknowledge to be human-aware indeed account for explicit human models, with most coming from those related to explanation or related literature \cite{ijcai2023p795,sreedharan2020expectation}.

Nevertheless, a closer look at all the papers revealed that many of them are built on top of assumptions that allow for the implicit modeling of humans. With this, in mind, we were able to easily categorize the assumptions into one of the dimensions discussed in the earlier section. We found that a vast majority of works focused on modeling knowledge state (particularly $\mathcal{M}_h$). {\em This shows a clear lack of work that focuses on $\mathcal{M}^R_h$, and as such are not as adaptive to the human's beliefs.} Moving away from the knowledge state, modeling of inferential capabilities and vocabulary was also considered by less works. The former could be explained by a general lack of robust tools to accurately capture and model human inferential capabilities. The most widely used model, i.e., noisy rational model \cite{noisyrat}, is known to be insufficient in many cases. In addition, we saw that there is less work overall in capturing vocabulary mismatch. This is particularly surprising given its prevalence within the larger XAI literature \cite{kim2018tcav}.

Moreover, we found that most works were focused on cases where the human assumes the role of the end user of the system. There were a few works that looked at humans as supervisors or teammates. However, this might be a result of the venue we chose or the keywords used. We expect to see more work if we had included robotic or multi-agent venues.

In regards to the inclusion of social science concepts and user studies, while there were works that addressed them, it was a clear minority. While most authors in the field publicly acknowledge the importance of both in works related to human-AI interaction, we see that in practice this is usually not the case.

\section{Conclusion and Future Directions}
In this survey paper, we hope to both provide a clear and concise description of what it means for an AI system to be considered human-aware. Starting with this description, we provide some characterization and properties of these models and then use it to perform an analysis of some recent works published in different prestigious AI conferences. In our analysis of the paper, we see many glaring omissions in terms of open problems and research opportunities. 
However, it is still worth noting that our paper focuses on a very small timeframe and only on four conferences. In the future, we hope to perform a more comprehensive survey that considers papers from a number of diverse venues over a larger timeframe.

\bibliography{ijcai19}

\end{document}